\documentclass{article}

\usepackage{arxiv}

\usepackage[utf8]{inputenc} 
\usepackage[T1]{fontenc}    
\usepackage{url}            
\usepackage{booktabs}       
\usepackage{amsfonts}       
\usepackage{nicefrac}       
\usepackage{microtype}      
\usepackage{lipsum}		
\usepackage{graphicx}
\usepackage{natbib}
\usepackage{doi}
\usepackage{amsmath}
\usepackage{algorithm}
\usepackage[noend]{algpseudocode}
\usepackage{multirow}
\hypersetup{hidelinks}
\usepackage{wrapfig}
\usepackage{svg}
\usepackage{tabularx}

\usepackage[compact]{titlesec}
\titlespacing\section{0pt}{1pt minus 1pt}{1pt minus 1pt}
\titlespacing\subsection{0pt}{1pt minus 1pt}{1pt minus 1pt}
\title{Solving the flexible job-shop scheduling problem through an enhanced deep reinforcement learning approach}

\author{Imanol Echeverria\thanks{Corresponding author.} \\
    TECNALIA, Basque Research and Technology Alliance (BRTA) \\
	Mikeletegi Pasealeakua 7, 20009 Donostia-San Sebastián (Spain) \\
	\texttt{imanol.echeverria@tecnalia.com} \\
	\And
	Maialen Murua\\
	TECNALIA, Basque Research and Technology Alliance (BRTA)\\
	Mikeletegi Pasealeakua 7, 20009 Donostia-San Sebastián (Spain) \\
	\texttt{maialen.murua@tecnalia.com} \\
 	\And
	Roberto Santana\\
	Computer Science and Artificial Intelligence Department\\
	University of the Basque Country\\
	Lardizabal Pasealeakua 1, 20018 Donostia-San Sebastián (Spain). \\
	\texttt{roberto.santana@ehu.eus} \\
}

\date{}

\renewcommand{\headeright}{}
\renewcommand{\undertitle}{}
\renewcommand{\shorttitle}{}

\hypersetup{
pdftitle={Solving the flexible job-shop scheduling problem through an enhanced deep reinforcement learning approach},
pdfsubject={cs.IA},
pdfauthor={Imanol Echeverria},
pdfkeywords={Solving the flexible job-shop scheduling problem through an enhanced deep reinforcement learning approach},
}

\begin{document}
\maketitle

\begin{abstract}
In scheduling problems common in the industry and various real-world scenarios, responding in real-time to disruptive events is essential. Recent methods propose the use of deep reinforcement learning (DRL) to learn policies capable of generating solutions under this constraint. The objective of this paper is to introduce a new DRL method for solving the flexible job-shop scheduling problem, particularly for large instances. The approach is based on the use of heterogeneous graph neural networks to a more informative graph representation of the problem. This novel modeling of the problem enhances the policy's ability to capture state information and improve its decision-making capacity. Additionally, we introduce two novel approaches to enhance the performance of the DRL approach: the first involves generating a diverse set of scheduling policies, while the second combines DRL with dispatching rules (DRs) constraining the action space. Experimental results on two public benchmarks show that our approach outperforms DRs and achieves superior results compared to three state-of-the-art DRL methods, particularly for large instances.
\end{abstract}

\section{Introduction}\label{sec:introduction}
Planning consists of assigning a series of operations to a set of resources within a specified time period to optimize one or multiple objectives. It is a widely studied problem in the literature due to its impact on the manufacturing industry \citep{gupta2006job} and various other real-world scenarios \citep{ccalics2015research}. One of the most well-known variants is the job-shop scheduling problem (JSSP) \citep{manne1960job}, in which a set of jobs, composed of operations, must be assigned to a set of machines. Each one of these operations can only be performed on a single given machine, and the order of the operations within a job must be respected. The objective of the problem is to create a resource allocation that optimizes one or multiple key performance indicators, with makespan being the most commonly used in the literature, defined as the time to complete all operations, as well as other metrics related to job tardiness or production costs such as energy consumption \citep{xiong2022survey}. Due to its complexity, this problem is known to be NP-Hard \citep{garey1976complexity}. A well-studied extension of the JSSP is the flexible job-shop scheduling problem (FJSSP), where the difference lies in the fact that each one of the operations can be processed on a set of machines instead of a single one. This problem is also NP-Hard as it adds the complexity of determining the optimal route for operations, resulting in a significantly larger search space of possible solutions. \\

Multiple approaches have been proposed in the literature to find optimal solutions for the FJSSP. Within the family of exact optimization methods, dynamic programming-based methods have been proposed \citep{kim2022dynamic}, although the most common ones are based on linear programming or mixed linear programming \citep{meng2020mixed, tutumlu2023mip}. However, due to the NP-hardness of the FJSSP, these methods cannot handle medium-large instances of the problem in a dynamic manner, that is, cannot generate optimal or pseudo-optimal solutions in real-time, and the sizes of the instances typically used for benchmarking are much smaller than those encountered in real-world scenarios. For example, \cite{meng2020mixed} used the first ten instances from the well-known Brandimarte benchmark \citep{brandimarte1993routing}, with the largest instance consisting of only twenty jobs and fifteen machines. This size is much smaller compared to real-world scenarios found in the industry, which can involve hundreds or thousands of jobs and dozens of machines \citep{da2019industrial}.\\

On the other hand, metaheuristic algorithms, particularly genetic algorithms \citep{gutierrez2011modular, chen2020self}, as well as approaches based on tabu search \citep{martinez2016moamp} or simulated annealing \citep{defersha2022mathematical}, have been widely employed. Although metaheuristic methods are more efficient than exact methods, both require a considerable amount of time to obtain satisfactory solutions as the algorithms need to iterate to find these solutions. Therefore, neither of these two approaches is suitable for environments with disruptive events, such as machine breakdowns or new jobs that need to be accommodated in real-time. Dispatching rules (DRs) \citep{calleja2014dispatching,chen2013flexible}, methods framed within the family of heuristics, are the most commonly used options in such situations. Each possible dispatching rule defines a particular a way to order the operations in a sequence, and a criterion to assign each operation to a machine. The DRs have the advantage of being capable of generating solutions quickly for large instances of the FJSSP. However, they present some disadvantages. Both criteria need to be manually designed by an expert, which requires time and effort, and they are also static, making them strategies that cannot adapt to diverse problem instances and cannot find global optimal solutions \citep{luo2021dynamic}. \\

Reinforcement learning (RL) and deep reinforcement learning (DRL) have emerged as promising alternatives to tackle problems that can be modeled as a Markov Decision Process (MDP), including other NP-Hard problems such as the Traveling Salesman Problem (TSP) \citep{cappart2021combining, mazyavkina2021reinforcement} or the Capacitated Vehicle Routing Problem (CVRP) \citep{kwon2020pomo}. Scheduling problems such as the FJSSP have received less attention in the literature but, in recent years, various DRL approaches have been proposed to tackle the JSSP and the FJSSP \citep{zhang2023deepmag, liu2023reinforcement}, with the former being more common as it is easier to model as an MDP. To apply RL to scheduling problems, the problem is first modeled as an MDP, and RL algorithms are used to learn a policy that sequentially assigns operations to machines while attempting to maximize the accumulated reward, which is usually designed to minimize the makespan. The advantage over metaheuristics is that, with each new instance, the model can reuse the knowledge gained from solving different instances.\\

The proposed DRL approaches outperform DRs, but there are still deficiencies that need to be addressed, particularly for large instances. The first deficiency is related to how the FJSSP is modeled as an MDP, specifically in the design of the state and action spaces. On the one hand, a state lacking properly modeled instance information (e.g., due to insufficient features) or containing unnecessary information hampers the policy's capacity for optimal decision-making. On the other hand, a high-dimensional action space makes exploration of the solution space more costly and model learning more complex. The second deficiency is related to how DRL models are used to generate solutions. A single solution can be produced using a greedy procedure, or multiple solutions can be generated through a process called sampling, which is more computationally costly. When using the greedy strategy, it is crucial that the policy acts robustly by not selecting actions leading to unsatisfactory results; however, current approaches do not constrain the action space. The sampling strategy has the drawback of generating excessively similar solutions since they are produced by the same model, thereby limiting the potential of this strategy.\\

To overcome these limitations, in this paper, we propose a novel DRL method to solve the FJSSP in real-time focusing on improvements in the modeling of the FJSSP as an MDP and proposing two approaches to enhance its performance. The contributions of this paper can be summarized as follows:

\begin{itemize}
\item A novel representation of the FJSSP as an MDP is presented, introducing a new job-type node and eliminating nodes and edges that are no longer relevant to the problem as the solution is constructed. This leads to a significantly reduced action space compared to other approaches and a state representation that facilitates the policy's information capture.
\item Limiting the action space of DRs to prevent the model to taking actions that are known to lead to poor results, producing a direct improvement of the final solutions, particularly in larger FJSSP instances. 
\item We propose an alternative to the sampling method in which solutions are generated using different models. To achieve this, we introduce a method for optimizing a hyperparameter domain to get a more diverse set of policies through Bayesian Optimization (BO) and a method for selecting the most suitable policies using the \textit{K}-Nearest Neighbors algorithm.
\item Multiple experiments have been conducted to validate this new approach. The method has been compared with DRs and three state-of-the-art DRL-based methods \citep{song2022flexible, wang2023flexible, lei2022multi} yielding improved results, particularly on larger instances on two public benchmarks. As far as we know, these three DRL approaches represent the state-of-the-art of DRL-based methods to solve the FJSSP.
\end{itemize}

The remainder of this paper is organized as follows. Section \ref{sec:relatedWork} presents recent research on the application of DRL methods to scheduling problems. In Section \ref{sec:preliminaries}, we introduce the FJSSP and GNNs. In Section \ref{sec:proposedmethod}, we describe how the FJSSP has been modeled as an MDP and the two methods to enhance the performance of DRL methods. In Section \ref{sec:experimentalresult}, we compare the proposed method with DRs and with three state-of-the-art DRL-based methods in two public benchmarks to verify the effectiveness of our approach. Finally, the conclusions and future work are presented in Section \ref{sec:conclusion}.

\section{Related work}\label{sec:relatedWork}
In this section, we will provide a brief analysis of the proposed methods for addressing the JSSP and the FJSSP using DRL-based methods to generate solutions in real-time. Firstly, we will present the contributions that use vectors and matrices to represent the state, and secondly, the ones that make use of graphs. \\

\cite{lin2019smart} proposed a framework for smart manufacturing based on a value-based RL method, specifically a deep Q-network (DQN), combined with edge computing to manage decisions across multiple edge devices. This framework encapsulates the state information into a feature vector that captures the relevant production state information, employing a multilayer perceptron (MLP) to determine the value of a state. Similarly, \cite{park2019reinforcement} solved the FJSSP for the semiconductor industry applying a DQN. \cite{luo2020dynamic} addressed the FJSSP with new job insertions with a DQN, which is employed to extract information from the production status using seven features that are used to select from six composite DRs. In their paper, \cite{luo2021dynamic} extended this work using the Double DQN algorithm (DDQN) to solve the multi-objective FJSSP, optimizing the total weighted tardiness and average machine utilization rate.\\

Instead of using vectors and value-based methods, \cite{liu2020actor} used convolutional neural networks (CNNs) and the deep deterministic policy gradient (DDPG) algorithm, which belongs to the policy-based family, was employed to tackle the JSSP. In this case, the input values are not a vector but a matrix with three channels: the first channel contains the process times of the operations, the second channel contains information about whether a job has been assigned to a machine, and the third channel indicates whether a job has been completed. This approach has the advantage of not having to design complicated features that reflect the production state, which can be time-consuming. Finally, \cite{zhang2023deepmag} proposed a multi-agent approach that combines DRL (specifically, DQNs) with Multi-Agent RL to solve the FJSSP. With this approach, each machine and job is considered an agent, and the different agents cooperate based on a multi-agent graph. \\

One significant limitation of many of these approaches is that they do not directly encapsulate much of the crucial information in the JSSP, particularly in the FJSSP, which resides in its structure and the relationships between operations and machines. Additionally, these methods solely rely on a greedy strategy for solution generation, leading to suboptimal results.\\

Building upon the work done in other domains such as vehicle routing \citep{tolstaya2021multi}, the information of a state can be represented using a graph, where nodes possess features that reflect their state and share information with connected nodes through edges. The state information can be extracted using GNNs and attention mechanisms \citep{vaswani2017attention, velivckovic2017graph}, which, unlike MLPs and CNNs, allow for capturing the structure of an instance and the relationships between nodes. Moreover, they are size-agnostic, meaning the network can adapt to a variable number nodes. One way to represent JSSP and FJSSP is through a disjunctive graph \citep{balas1969machine}, where nodes represent operations and two types of edges are employed. Directed edges indicate precedence among operations belonging to the same job, while undirected edges connect pairs of operations that can be performed on the same machine.\\

Using this representation, \cite{zhang2020learning} proposed a method to learn DRs using an end-to-end DRL agent for the JSSP, employing the Proximal Policy Optimization (PPO) \citep{schulman2017proximal} algorithm to train the policy. Using the disjunctive graph representation, the authors in \cite{lei2022multi} proposed to model the FJSSP as multiple MDPs and employed a multi-pointer graph networks, specifically a graph isomorphism network (GIN), to learn two sub-policies applied sequentially: first, one for operation selection and, second, another for machine assignment. However, this approach may pose a problem since the selection of the machine should be as important as the operation, and the choice of the former is contingent on the latter. \cite{song2022flexible} presented the use of heterogeneous graphs, modeling the state using operation-type and machine-type nodes, connecting operations with machines through edges, and heterogeneous graph attention networks (HGATs) to extract the features from a state. This representation marks a significant improvement compared to the disjunctive graph, as it substantially decreases the number of edges, thereby reducing the size of the action space. Furthermore, building upon other approaches proposed for problems such as the TSP or CVRP \citep{kwon2020pomo}, they proposed the use of the sampling strategy for the FJSSP, generating a set of 100 solutions. This approach led to improved results despite incurring in a higher computational cost. Based on this work, \cite{wang2023flexible} improved the attention models for deep feature extraction, employing a dual-attention network with operation message attention blocks and machine message attention blocks. In this work, they explore the importance of modeling the FJSSP as an MDP to obtain a more precise state representation by investigating the significance of the diverse relationships between operations and machines, for which they use attention mechanisms. However, there is still room for enhancements in attention models to more accurately capture the relationships between different types of nodes. \\

These works represent advancements in how the FJSSP is modeled as an MDP, but there are still challenges that remain unresolved, as this field is still relatively unexplored. In particular, these works do not consider the job-type nodes, only operations and machines, as part of the state space, even though it is a fundamental aspect of the problem. The incorporation of this entity may simplify policy learning and lead to improved results. Another aspect to achieve this goal, not adequately addressed in these works, is the elimination of unnecessary information, such as operations that have already been completed. An additional issue related to the action space is that as the problem instance size increases, the dimensionality grows significantly, especially if it is defined using the machine and operations nodes. This high dimensionality can lead to inefficient exploration of the solution space.\\

Finally, as highlighted earlier, employing a sampling strategy can enhance the outcomes achievable with DRL methods, though this advantage incurs in a higher computational cost. The greedy strategy involves selecting an action based on the probability distribution defined by the policy, rather than always choosing the most rewarding action (as in a greedy strategy). Using the sampling strategy a set of solutions are generated instead of one, with the final solution being the one that yields the best score. However, this method may face the challenge of generating too similar solutions, as they are derived from the same policy distribution. This similarity can result in only marginal improvements in the overall results, limiting the efficacy of the approach. \\

Table \ref{tab:related} summarizes recent works for solving the FJSSP in real-time. In the table, the RL algorithm, the type of neural network used for feature extraction, how the state and action spaces were modeled, and how the model was employed to generate solutions have been highlighted. Regarding the state and action spaces, the last four approaches use graphs to represent states, and the sets $\mathcal{O}, \mathcal{M}$ and $\mathcal{J}$ indicate the use of nodes of operation, machine, and job types, respectively. In the table, it can be observed how early works utilized value-based RL algorithms and MLPs, while more recent ones tend to employ policy methods and graphs to model the problem.

\begin{table}[H]
\caption{Survey of DRL-methods for solving the FJSSP in real-time.}
\label{tab:related}
\resizebox{\textwidth}{!}{
\begin{tabular}{l r r r r r r}
        \hline
        Method & RL Algorithm & Feature extractor & State space & Action space & Inference strategy \\
        \hline
        \hline
         \cite{park2019reinforcement}&DQN& MLP &State features& Operations & Greedy\\
         \cite{luo2020dynamic}&DQN& MLP &State features& DR selection & Greedy\\ 
         \cite{luo2021dynamic}&DDQN& MLP &State features & DR selection & Greedy\\ 
         \cite{zhang2023deepmag}&DQN&  MLP & State features  & Job-Machines & Greedy\\  
         \cite{lei2022multi}&PPO& GIN & $\mathcal{O}^*$, Machine features & $\mathcal{O} \times \mathcal{M}$ & Greedy\\ 
         \cite{song2022flexible}&PPO&  HGAT & $\mathcal{O}, \mathcal{M}$ & $\mathcal{O} \times \mathcal{M}$ & Greedy, Sample \\ 
         \cite{wang2023flexible}&PPO&  HGAT & $\mathcal{O}, \mathcal{M}$ & $\mathcal{O} \times \mathcal{M}$ &Greedy, Sample \\ 
         \textbf{Ours}&PPO&HGATv2$^{**}$ & $\mathcal{O}, \mathcal{M}, \mathcal{J}$ & Masked $\mathcal{J} \times \mathcal{M}$& Greedy, Diverse models\\
         \hline
    \end{tabular}}
\footnotesize{$^*$ The set $\mathcal{O}$ indicates that only operation-type nodes are used. The sets $\mathcal{M}$ and $\mathcal{J}$ refer to machine-type and job-type nodes, respectively.}\\
\footnotesize{$^{**}$ HGATv2 is a type of HGAT that utilizes a more expressive attention mechanism, which will be explained in the following section.}
\end{table}

\section{Preliminaries}\label{sec:preliminaries}
In this section, we will briefly describe how the FJSSP is formulated, as well as introducing key concepts of the GNNs that will be used in our proposed method.

\subsection{Problem formulation}
An instance of the FJSSP problem is defined by a set of jobs $\mathcal{J} = \{j_1 , j_2, \ldots, j_n \}$, where each $j_i \in \mathcal{J}$ is composed of a set of operations $\mathcal{O}_{j_i} = \{ o_{i1}, o_{i2}, \ldots o_{im}\}$, and each operation can be performed on one or more machines from the set $\mathcal{M} = \{m_1, m_2, \ldots , m_p\}$. The processing time of operation $o_{ij}$ on machine $m_{k}$ is defined as $p_{ijk} \in \mathbb{R}^+$. We define $\mathcal{M}_{o_{ij}} \subseteq \mathcal{M}$ as the subset of machines on which that operation $o_{ij}$ can be processed, $\mathcal{O}_{j_i} \subseteq \mathcal{O}$ as the set of operations that belong to the job $j_i$ where $\bigcup\limits_{i = 1}^n \mathcal{O}_{j_i} = \mathcal{O}$, and $\mathcal{O}_{m_k} \subseteq \mathcal{O}$ as the set of operations that can be performed on machine $m_k$. The execution of operations on machines must satisfy a series of constraints:
\begin{itemize}
\item All machines and jobs are available at time zero.
\item A machine can only execute one operation at a time, and the execution cannot be interrupted.
\item An operation can only be performed on one machine at a time.
\item The execution order of the set of operations $\mathcal{O}_{j_i}$ for every $j_i \in \mathcal{J}$ must be respected.
\item Job executions are independent of each other, meaning no operation from any job precedes or has priority over the operation of another job.
\end{itemize}
In essence, the FJSSP combines two problems: a machine selection problem, where the most suitable machine is chosen for each operation, a routing problem, and a sequencing or scheduling problem, where the sequence of operations on a machine needs to be determined. Given an assignment of operations to machines, the completion time of a job, $j_i$, is defined as $C_{j_i}$, and the makespan of a schedule is defined as $C_{max} = \max\limits_{j_i \in \mathcal{J}} C_{j_i}$, which is the most common objective to minimize.

\subsection{Graph Neural Networks} \label{sssec:GNN}

GNNs \citep{scarselli2008graph} have received significant attention in recent years \citep{zhang2021graph, zhou2020graph} due to their ability to efficiently extract information from structured data in the form of graphs, where predictions are made based on the edges connecting the nodes. A directed graph $\mathcal{G}$ is defined as a tuple of two sets $(\mathcal{V}, \mathcal{E})$ where $\mathcal{V}$ contains nodes, $\mathcal{V} = \{1, \ldots, n\}$,  and $\mathcal{E}$ edges, $\mathcal{E} \subseteq \mathcal{V} \times \mathcal{V}$, where $(i, j) \in E$ represents an edge from a node $i$ to a node $j$. It is assumed that every node has an initial representation $\boldsymbol{h_i}^{(0)} \in \mathbb{R}^{d_0}$, where $d_0$ is the number of the initial features of the node,  and that all edges are directed, so an undirected graph can be represented with bidirectional edges. Multiple GNN architectures have been proposed \citep{wu2020comprehensive}, but one of the most commonly used ones is the Graph Attention Network (GAT) \citep{velivckovic2017graph}, where each node updates its representation by aggregating the representations of its neighbors. Specifically, given a set of node representations $\mathcal{H} = \{ \boldsymbol{h}_1 , \boldsymbol{h}_2 \ldots \boldsymbol{h}_n\}$, the update of the representation of a node $\boldsymbol{h}_i^\prime$ is performed as follows:
\begin{equation}
\boldsymbol{h_i}^{\prime}=f_\theta\left(\boldsymbol{h}_i, \text{AGGREGATE}\left(\left\{\boldsymbol{h}_j \mid j \in \mathcal{N}_i\right\}\right)\right)
\end{equation}
where $\mathcal{N}_i$ represents the set of nodes connected to node $i$, while $f$ and $\text{AGGREGATE}$ are functions that define the GNN architecture. Many GNNs assign equal importance to all neighboring nodes, but in \cite{velivckovic2017graph}, the authors proposed a weighted average of the representations of $\mathcal{N}_i$ through a scoring function that learns the importance of node $j$'s features to node $i$. The scoring function is defined as:
\begin{equation}
e\left(\boldsymbol{h}_i, \boldsymbol{h}_j\right)=\text { LeakyReLU }\left(\boldsymbol{a}^{\top} \cdot\left[\boldsymbol{W} \boldsymbol{h}_i \| \boldsymbol{W} \boldsymbol{h}_j\right]\right)
\end{equation}
where $\boldsymbol{a} \in \mathbb{R}^{2d^\prime}$ and $\boldsymbol{W} \in \mathbb{R}^{d^\prime \times d_0}$  are learned vectors, $d^\prime$ is the length of the vector $\boldsymbol{h_i}^\prime$  and $||$ denotes vector concatenation. These scores are then normalized using the softmax function across all neighbors of the node, resulting in:
\begin{equation}
\alpha_{i j}=\operatorname{softmax}_j\left(e\left(\boldsymbol{h}_i, \boldsymbol{h}_j\right)\right)=\frac{\exp \left(e\left(\boldsymbol{h}_i, \boldsymbol{h}_j\right)\right)}{\sum_{j^{\prime} \in \mathcal{N}_i} \exp \left(e\left(\boldsymbol{h}_i, \boldsymbol{h}_{j^{\prime}}\right)\right)}
\end{equation}
Finally, to obtain $\boldsymbol{h}_i^{\prime}$, a weighted sum of the representation of neighboring nodes is computed using the attention coefficients, followed by a non-linear activation function $\sigma$:
\begin{equation}
\boldsymbol{h}_i^{\prime}=\sigma\left(\sum_{j \in \mathcal{N}_{i}} \alpha_{i j} \cdot \boldsymbol{W} \boldsymbol{h}_j\right)
\end{equation}
\cite{brody2021attentive} proposed GATv2, which introduces a change in the internal order of these operations to achieve a more expressive attention mechanism. The problem with the original order is that the vectors $\boldsymbol{W}$ and $\boldsymbol{a}$ are applied consecutively, leading to them being collapsed into a single linear layer. The proposed solution is to apply $\boldsymbol{a}$ after the non-linear function and $\boldsymbol{W}$ after the concatenation, resulting in a more expressive score for each node-key pair. The updated formula becomes:
\begin{equation}
e\left(\boldsymbol{h}_i, \boldsymbol{h}_j\right)=\boldsymbol{a}^{\top} \operatorname{LeakyReLU}\left(\boldsymbol{W} \cdot\left[\boldsymbol{h}_i \| \boldsymbol{h}_j\right]\right)
\end{equation}
and $\boldsymbol{W}$ can be divided into $[\boldsymbol{W_1} \| \boldsymbol{W_2}] | \boldsymbol{W_1}, \boldsymbol{W_2}  \in \mathbb{R}^{d^\prime \times d}$ where each matrix represents the left half and right half of the columns of $\boldsymbol{W}$.

\section{Proposed method}\label{sec:proposedmethod}
In this section, we will describe the proposed methodology to obtain real-time solutions for the FJSSP. Our contribution lies in two key areas: the modeling of the FJSSP as an MDP, and the implementation of two methods to enhance the performance of the model trained through DRL. These methods include constraining the action space using DRs and creating a set of diverse policies. What follows is a summary of the methodology.\\

\begin{figure}[htbp] 
	\centering 
	\includegraphics[width=1\columnwidth]{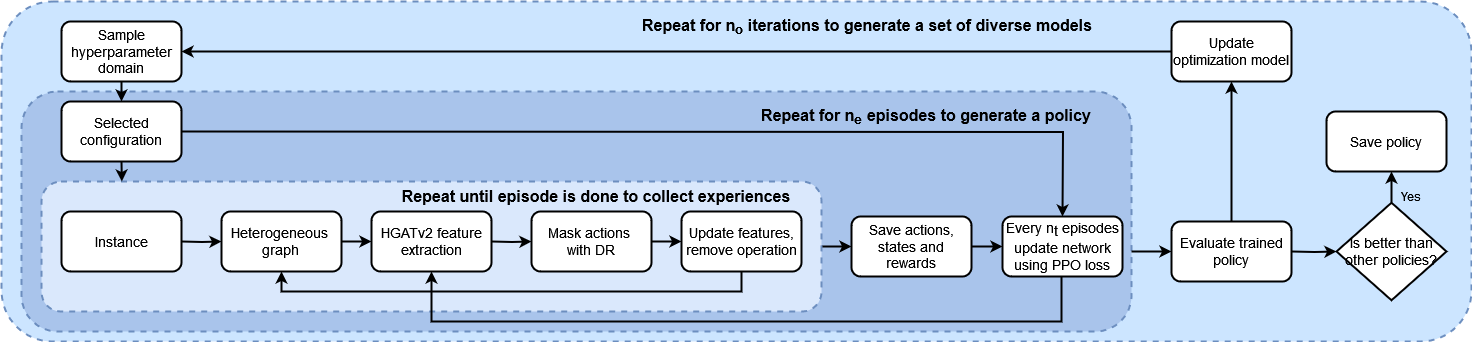} 
	\caption{Diagram summarizing the different components of the proposed method.} 
	\label{fig:all_method} 
\end{figure}

Our methodology is divided into two phases: the training phase and the inference phase. During the training phase, a set of diverse policies is generated. In the inference phase, these policies are tested on new instances. Figure~\ref{fig:all_method} illustrates the steps involved in generating a set of diverse policies, organized into three blocks. From the largest to the smallest block, the first one involves selecting configurations to generate a policy; the second describes the process of training a policy; and the smallest block details the method for solving an instance. Two distinct strategies can be utilized during the inference phase: the first strategy generates a single solution using one policy, while the second employs a representative subset of models in parallel to produce multiple solutions. The training phase consists of the following steps:

\begin{itemize}
    \item \textbf{Validation set creation}: We begin by creating a validation set. These instances are optimally solved using CP, providing a baseline against which the performance of subsequently generated models is evaluated.
    \item \textbf{Hyperparameter domain}: A domain of hyperparameters is established to guide policy generation. This includes the creation of instances biasing its characteristics, such as the number of operations, machines, processing times, etc., the configuration of the diverse GNN architecture, and different parameterizations of the training process, among other elements. For example, this results in each policy being trained on a unique set of instances, differing in distribution from the other sets.
    \item \textbf{Creation of candidate set}: A set of candidate policies is also created to be used in inference. Initially, this set is populated with a policy generated through sampling the hyperparameter domain, providing a baseline for evaluating the performance of the following generated policies.
    \item \textbf{Objective and optimization approach}: Our goal is to create a diverse set of policies, each tailored for different instances, ensuring thorough coverage during inference. To achieve this, we use an optimization strategy focused on refining policy generation. We aim to identify policies that perform better than those currently in the candidate set. This approach helps to maintain a robust and adaptable policy suite, equipped for various scheduling scenarios.
    \item \textbf{Iteration process}: Within each optimization cycle, the hyperparameter domain is sampled to train a  policy over $n_e$ episodes. In each episode, an instance is created and transformed into a heterogeneous graph, which is processed by an HGNN (specifically, HGATv2). This forms a probability distribution over potential actions (job assignments to machines). These distributions are then sampled—guided by a DR—to gather varied training samples for training. The HGNN is updated every $n_t$ episodes using the PPO loss function, based on accumulated training samples.
    \item \textbf{Policy evaluation}: After $n_e$ episodes, the new policy is tested against the validation set. Performance is measured by comparing the gap to CP-derived optimal results. Policies that improve upon the results of the current candidate set are included in this set, and the policy generation model is updated accordingly.
\end{itemize}

In the next subsections, a more detailed version of these steps will be provided. First, we will explain how we have formulated the FJSSP as an MDP, defining the state space, action space, reward function, and state transitions. Secondly, we will present how we have modeled the policy using a GNN based on attention mechanisms and message passing between different nodes. Thirdly, we will describe two methods that are part of our DRL approach to enhance its performance, particularly in large instances, by applying masks based on DRs and using a novel method for generating a diverse set of scheduling policies.

\subsection{The FJSSP as a Markov Decision Process.} 

To define an MDP, we need to specify the space of states, the action space, the reward and the state transitions. In broad terms, at each step, when an operation is completed, the agent must decide which operation to perform next and in which machine.\\

\textbf{State space.} Each state, which at step $t$ is denoted as $s_t$, is a heterogeneous graph $\mathcal{G}_t=(\mathcal{V}_t, \mathcal{E}_t)$, where the set $\mathcal{V}_t$ consists of three types of nodes: the set of operations $\mathcal{O}_t$, the set of machines $\mathcal{M}_t$, and the set of jobs $\mathcal{J}_t$. Unlike previous works, the set of operations only includes those that have not yet been scheduled. Each of these types of nodes has the following features: 
\begin{itemize} 
\item The operation nodes have a Boolean feature indicating whether the operation can be performed or not and the pending processing time. For an operation node, the second feature is calculated as the sum of the average processing times of the remaining operations. 
\item The machine nodes have two features: the time at which the last assigned operation on the machine is completed and the utilization percentage of that machine. 
\item The features of job nodes contain most of the information about the state of the instance. Their features include whether the job has been completed, the time at which its last assigned operation has been finished, the number of operations remaining for that job and the pending processing time. \end{itemize}
The set of edges $\mathcal{E}_t$ consists of the following types of edges: 
\begin{itemize} 
\item Undirected edges between machines and operations, connecting operations that can be executed in machines and indicating their processing time. An undirected edge can be represented with two sets of directed edges, one between the operations and the machines, $\mathcal{OM}_t \subseteq \mathcal{O}_t \times \mathcal{M}_t$, and another between the machines and the operations, $\mathcal{MO}_t  \subseteq \mathcal{M}_t \times \mathcal{O}_t$. Both have features which are the processing time, the ratio of the processing time to the maximum processing time of the operation, and the ratio to the maximum of all the options that the machine can process.
\item Directed edges between operations and jobs, indicating that an operation belongs to a job, $\mathcal{OJ}_t  \subseteq \mathcal{O}_t \times \mathcal{J}_t$.
\item Directed edges between operations, indicating that an operation immediately precedes the other, $\mathcal{OO}_t  \subseteq \mathcal{O}_t \times \mathcal{O}_t$. 
\item Directed edges between machines and jobs, adding the processing time and the gap that would be generated by executing that job (i.e., the first operation that can be performed for that job) on that machine. It also has as features the same ratios as the edges between the operations and machines. The set of edges is defined as $\mathcal{MJ}_t  \subseteq \mathcal{M}_t \times \mathcal{J}_t$ and each edge $(m_k, j_i) \in \mathcal{MJ}_t$ has features defined as $pg_{ki}$.
\item Jobs are connected to each other to enable information communication among their states, $\mathcal{JJ}_t  \subseteq \mathcal{J}_t \times \mathcal{J}_t$.
\item Machines are connected to each other to enable information communication among their states,  $\mathcal{MM}_t  \subseteq \mathcal{M}_t \times \mathcal{M}_t$.
\end{itemize}

It should be noted that in our case, the number of features associated with nodes is reduced in comparison to previous proposals in the literature, resulting in a lower computational cost as less information needs to be calculated at each state. In comparison, in \cite{song2022flexible}, operations have six features and machines have three features and in \cite{wang2023flexible} operations have ten features, machines eight, and compatible operation-machine pairs also eight features.  \\ 

\textbf{Action space.} At each step $t$, the set of actions $\mathcal{A}_t$ is composed of all pairs of compatible jobs and machines. When choosing a job, the first operation of the job that has not yet been scheduled is selected. This set corresponds to the set of edges that connect jobs and machines, $\mathcal{MJ}_t$. As explained in the next section, to improve learning efficiency and stabilize inference, based on a DR, some of these actions will be excluded from being eligible by the policy. \\

The inclusion of a job node offers two main advantages. Firstly, it significantly reduces the action space, as the number of jobs is usually much smaller than the number of operations. Secondly, it concentrates the relevant information of the problem on a single type of node, making the learning process more efficient and the inference faster. \\

\textbf{State Transitions.} Performing an action $a_t$ in state $s_t$ transitions to a new state $s_{t+1}$ where the operation has been executed. During each state transition, the heterogeneous graph undergoes three transformations. Firstly, the operation node that has been executed is removed, along with all the edges connected to it. Secondly, the edges of the executed job are removed, and new edges are created with the next operation that needs to be performed. Lastly, the features of all nodes and edges are updated. \\

\begin{figure}[!h] 
	\centering 
	\includegraphics[width=1\columnwidth]{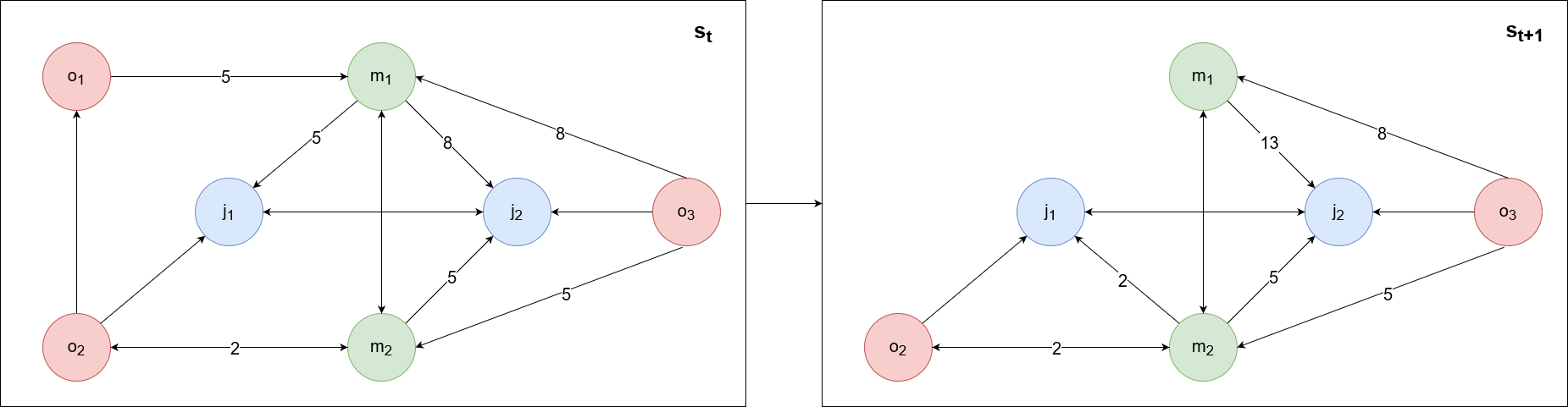} 
	\caption{In the top figure, a simple example of a FJSSP instance is represented using our proposed representation. In the bottom figure, $o_1$ has been assigned to $m_1$ and a node and some edges of the graph have been removed.} 
	\label{fig:graphs} 
\end{figure}

In Figure \ref{fig:graphs}, an example of an instance of the FJSSP, and the transition between two states is shown. Blue nodes represent jobs, red nodes represent operations, and green nodes represent machines. The edges are directed, and the values on the edges represent their processing times. In this instance, there are two jobs, $j_1$ composed of $o_1$ and $o_2$, and $j_2$ composed of $o_3$. $o_1$ can be processed on machine $m_1$ in five time units, $o_2$ on machine $m_2$ in three time units, and $o_3$ on $m_1$ in eight units and on $m_2$ in five units. The action taken is to select the edge $(m_1, j_1)$, which corresponds to assigning operation $o_1$ to machine $m_1$. After this action, several steps are performed. First, the node $o_1$ and all its edges are removed. Then, the edges of node $j_1$ are updated to correspond to its next operation, $o_2$. Finally, the edge of $j_2$ connected to $m_1$ is updated because performing an operation on this machine now requires waiting until $o_1$ is completed.\\

\textbf{Reward function.} The objective of the agent is to minimize the makespan. As proposed in multiple papers \citep{zhang2020learning, lei2022multi}, one way to minimize the total makespan is by defining the reward function at step $t$ as $r(s_t, a_t, s_{t+1}) = C(s_t) - C(s_{t+1})$, where $C(s_t)$ represents the makespan at step $t$. If the discount factor, $\gamma$ is set to 1, the accumulated reward becomes $\sum_{t=0}^{|\mathcal{O}|} r\left(s_t, a_t, s_{t+1}\right) = C\left(s_0\right) - C\left(s_{|\mathcal{O}|}\right)$. Assuming that the makespan is zero or a constant in the first step, minimizing the accumulated reward is equivalent to minimizing $C_{s_{|\mathcal{O}|}}$, which is the makespan in the final state. 

\subsection{Feature extraction}

To extract the state information, which will be used to generate a policy, we employed a HGNN that utilizes the attention mechanisms of GATv2 explained in Section \ref{sssec:GNN}. This approach differs from the one used by \cite{song2022flexible} and  \cite{wang2023flexible}, where they employ a simpler attention mechanism. As demonstrated by \cite{brody2021attentive}, it is possible to obtain GNN architectures with higher and more dynamic attention capacity without significantly increasing the computational cost. This is crucial for a problem such as the FJSSP. For each type of node, the initial representations are denoted as follows: $\boldsymbol{h_{j_i}} \in \mathbb{R}^{d_{\mathcal{J}}}$ for jobs, $\boldsymbol{h_{o_{ij}}} \in \mathbb{R}^{d_{\mathcal{O}}}$ for operations, and $\boldsymbol{h_{m_i}} \in \mathbb{R}^{d_{\mathcal{M}}}$ for machines, where $d_{\mathcal{O}}, d_{\mathcal{J}}, d_{\mathcal{M}}$ are the number of initial features of operations, jobs and machines respectively. The following section describes how the embeddings of the three types of nodes that compose the state are calculated. \\

\textbf{Operation embeddings.} Each operation receives information from its successor and the machines on which it can be processed. Therefore, it is necessary to calculate two attention weights. For each $o_{ij} \in {\mathcal{O}_{j_i}}_t$ with $j \leq | {\mathcal{O}_{j_i}}_t| - 1$, the weight $e_{o_{ijj+1}}$ between the preceding operations $o_{ij}$ and $o_{ij+1}$ is calculated as:
\begin{equation}
e_{o_{ijj+1}}= {\boldsymbol{a^{\mathcal{O}}}}^{\top} \text { LeakyReLU }\left(\boldsymbol{W_1^\mathcal{O}} \boldsymbol{h}_{o_{ij}} + \boldsymbol{W_2^\mathcal{O}}\boldsymbol{h}_{o_{ij+1}}\right)
\end{equation}
where, ${\boldsymbol{a^{\mathcal{O}}}}^{\top} \in \mathbb{R}^{2d_{\mathcal{O}}^\prime }$ and $\boldsymbol{W_1^{\mathcal{O}}}, \boldsymbol{W_2^{\mathcal{O}}} \in \mathcal{R}^{d_{\mathcal{O}}^\prime \times d_{\mathcal{O}}}$ are learned linear transformations for all $j \leq |{\mathcal{O}_{j_i}}_t| - 1$ and $d_{\mathcal{O}}^\prime$ is the size of the dimension of the embedding of the operation. We also calculate the attention coefficient between a operation with itself, $e_{o_{ijj}}$. To calculate the attention between the nodes of machines and operations, considering that these edges also have features, we use $e_{om_{ijk}}$ to represent the edge connecting operation $o_{ij}$ and machine $m_k$, and $\boldsymbol{h_{p_{ijk}}} \in \mathbb{R}^{d_{\mathcal{OM}}}$ as the initial representation of the features of that edge. The weight between nodes $o_{ij}$ and $m_k$ is  calculated as follows:
\begin{equation}
e_{om_{ijk}}={\boldsymbol{a^{\mathcal{OM}}}}^{\top}  \text { LeakyReLU }\left(\boldsymbol{W_1^\mathcal{OM}} \boldsymbol{h}_{o_{ij}}  +  \boldsymbol{W_2^\mathcal{OM}}\boldsymbol{h}_{m_k} + \boldsymbol{W_3^\mathcal{OM}}\boldsymbol{h_{p_{ijk}}}\right)
\end{equation}
where, ${\boldsymbol{a^{\mathcal{OM}}}}^{\top} \in \mathbb{R}^{3d_{\mathcal{OM}}^\prime}$ and $\boldsymbol{W_1^{\mathcal{OM}}} \in \mathcal{R}^{d_{\mathcal{OM}}^\prime \times d_{\mathcal{O}}} , \boldsymbol{W_2^{\mathcal{OM}}} \in \mathcal{R}^{d_{\mathcal{OM}}^\prime \times d_{\mathcal{M}}} , \boldsymbol{W_3^{\mathcal{OM}}} \in \mathcal{R}^{d_{\mathcal{OM}}^\prime \times d_{\mathcal{OM}}}$. After normalizing the attention weights that the operation has with its successor and itself, and the weight between the operation and the machines using a softmax function, the normalized attention coefficients $\alpha_{o_{ijj}}$, $\alpha_{o_{ijj+1}}$, and $\alpha_{om_{ijk}}$ are obtained, and the new representation of the operation $\boldsymbol{h_{o_{ij}}}^\prime$ is calculated as follows:
\begin{equation}
\boldsymbol{h_{o_{ij}}}^\prime= \text{ELU}\left(\alpha_{o_{ijj}} \boldsymbol{W_2^\mathcal{O}} \boldsymbol{h_{o_{ij}}} + \alpha_{o_{ijj+1}} \boldsymbol{W_2^\mathcal{O}} \boldsymbol{h_{o_{ij+1}}} + \sum_{m_k  \in \mathcal{M}_{o_{ij}} }\alpha_{om_{ijk}}\left(\boldsymbol{W_2^\mathcal{OM}}\boldsymbol{h_{m_k}} + \boldsymbol{W_3^\mathcal{OM}}\boldsymbol{h_{p_{ijk}}}\right)\right)
\end{equation}
\textbf{Machine embeddings.} Machines have two connections from which they will receive information, operations and other machines. The purpose of machines communicating with each other is to be able to understand their state and prioritize their tasks accordingly. The first weight is calculated similarly to $e_{om_{ijk}}$, and it is denoted as $e_{mo_{ijk}}$. To calculate the attention that a machine should assign to other machines, for all $m_i, m_j \in \mathcal{M}$, we compute:
\begin{equation}
e_{m_{ij}}= {\boldsymbol{a^{\mathcal{M}}}}^{\top} \text { LeakyReLU }\left(\boldsymbol{W_1^\mathcal{M}} \boldsymbol{h_{m_{i}}} + \boldsymbol{W_2^\mathcal{M}}\boldsymbol{h_{m_{j}}}\right)
\end{equation}
where, ${\boldsymbol{a^{\mathcal{M}}}}^{\top} \in \mathbb{R}^{2d_{\mathcal{M}}^\prime }$ and $\boldsymbol{W_1^{\mathcal{M}}}, \boldsymbol{W_2^{\mathcal{M}}} \in \mathcal{R}^{d_{\mathcal{M}}^\prime \times d_{\mathcal{M}}}$. After normalizing the attention coefficients, $\boldsymbol{h_{m_{i}}}^\prime$ is calculated as:
\begin{equation}
\boldsymbol{h_{m_{i}}}^\prime= \text{ELU}\left(\sum_{m_k  \in \mathcal{M} }\alpha_{m_{ik}}\boldsymbol{W_2^\mathcal{M}}\boldsymbol{h_{m_k}} + \sum_{o_{ij}  \in \mathcal{O}_{m_{i}} }\alpha_{mo_{ijk}}\left(\boldsymbol{W_2^\mathcal{MO}}\boldsymbol{h_{o_{ij}}} + \boldsymbol{W_3^\mathcal{MO}}\boldsymbol{h_{p_{ijk}}}\right)\right)
\end{equation}
where, $\boldsymbol{W_2^{\mathcal{MO}}} \in \mathcal{R}^{d_{\mathcal{O}}^\prime \times d_{\mathcal{O}}} , \boldsymbol{W_3^{\mathcal{MO}}} \in \mathcal{R}^{d_{\mathcal{MO}}^\prime \times d_{\mathcal{MO}}}$. \\

\textbf{Job embeddings.} The calculation of the embeddings for jobs has three sources of information: the information coming from the operations, the machines, and the other jobs. We define ${\mathcal{MJ}_{j_i}}_t$ as the set of machines that can be assigned the job $j_i$ at a step $t$ and $\boldsymbol{h_{pg_{ki}}} \in \mathbb{R}^{d_{\mathcal{MJ}}}$ as the initial representation of the features of that edge. In summary, by calculating the attention coefficients in a similar manner to the previously described embeddings, $\boldsymbol{h}_{j_{i}}^\prime$ is calculated as

\begin{equation}
\boldsymbol{h_{j_{i}}}^\prime= \text{ELU}\left(\sum_{j_j  \in \mathcal{J} }\alpha_{j_{ij}}\boldsymbol{W_2^\mathcal{J}}\boldsymbol{h_{j_j}} + \sum_{o_{ij}  \in {\mathcal{O}_{j_i}}_t }\alpha_{jo_{ij}}\boldsymbol{W_2^\mathcal{JO}}\boldsymbol{h_{o_{ij}}}  + \sum_{m_{k}  \in {\mathcal{MJ}_{j_i}}_t }\alpha_{mj_{ki}}\left(\boldsymbol{W_2^\mathcal{MJ}}\boldsymbol{h_{m_{k}}} + \boldsymbol{W_3^\mathcal{MJ}}\boldsymbol{h_{pg_{ki}}}\right)\right)
\end{equation}
where, $\boldsymbol{W_2^{\mathcal{J}}} \in \mathcal{R}^{d_{\mathcal{J}}^\prime \times d_{\mathcal{J}}} , \boldsymbol{W_2^{\mathcal{JO}}} \in \mathcal{R}^{d_{\mathcal{O}}^\prime \times d_{\mathcal{O}}} , \boldsymbol{W_2^{\mathcal{MJ}}} \in \mathcal{R}^{d_{\mathcal{M}}^\prime \times d_{\mathcal{M}}} , \boldsymbol{W_3^{\mathcal{MJ}}} \in \mathcal{R}^{d_{\mathcal{MJ}}^\prime \times d_{\mathcal{MJ}}}$. \\

The process just described details how to perform the calculation for the first layer of the GNN. However, this process can be repeated \emph{L} times, generating the embeddings ${\boldsymbol{h_{j_{i}}}^\prime}^{(L)}$, ${\boldsymbol{h_{o_{ij}}}^\prime}^{(L)}$, ${\boldsymbol{h_{m_{k}}}^\prime}^{(L)}$, where each layer uses the embeddings of jobs, operations, and machines from the previous layer, and all layers make use of $\boldsymbol{h_{pg_{ki}}}$ and $\boldsymbol{h_{p_{ijk}}}$, i.e., the original edge features.

\subsection{Generating a policy}

To generate a policy, we employed the well-known PPO algorithm \citep{schulman2017proximal}. This involves generating a policy (actor) $\pi_\theta(a_t | s_t)$ and a value network (critic) $v_\theta(s_t)$. Both networks use the embeddings calculated by the GNN, although in different ways. For the policy, we make predictions using the embeddings of the edges $\mathcal{MJ}_t$. For each $(m_k, j_i) \in \mathcal{MJ}_t$, we concatenate the embeddings of the machine ${\boldsymbol{h_{m_{k}}}^\prime}^{(L)}$, the job ${\boldsymbol{h_{j_{i}}}^\prime}^{(L)}$, and the edge feature $\boldsymbol{h_{pg_{ki}}}$, and then apply $\text{MLP}_{1\theta}$, a mask (based on DRs), that will be presented in the next section, and a softmax function to create a probability distribution over all possible pairs of jobs and machines. For the critic, we apply another $\text{MLP}_{2\theta}$ to all job embeddings, with hyperbolic tangent as the activation function, and perform mean pooling to obtain a single value, as described in Equation \ref{eq:mlp} :

\begin{equation}v_\theta(s_t) = \frac{1}{|\mathcal{J}_t|}\sum_{j_i \in \mathcal{J}_t}\text{MLP}_{2\theta}\left({\boldsymbol{h_{j_{i}}}^\prime}^{(L)}\right)\label{eq:mlp}\end{equation}

\begin{figure}[htbp] 
	\centering 
	\includegraphics[width=1\columnwidth]{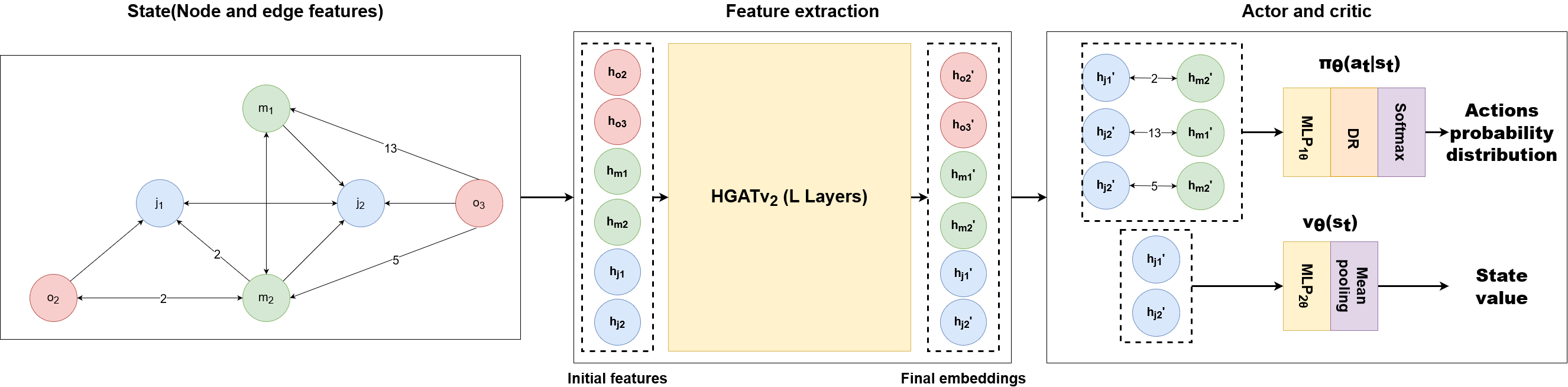} 
	\caption{The network architecture of our approach, EDSP.} 
	\label{fig:archs} 
\end{figure}

To train both networks, we run the policy for $n_{eps}$ episodes, where in each episode, the agent interacts with the environment by sampling from the probability distribution $\pi_\theta(a_t | s_t)$, and the experiences are stored in a buffer. A schematic representation of the architecture of our approach is presented in Figure \ref{fig:archs}. Every $n_t$ episodes, we use these experiences to train the networks for $K$ epochs in batches of size $b$. Additionally, every $n_g$ episodes, a new set of synthetic instances of dimension $n_{ins}$ is generated for the agent to interact with.

\subsection{Enhancing the performance of the DRL method}

While the method introduced in the previous section can generate good quality solutions, in this paper we introduce two strategies to enhance its performance. The methods are validated using our algorithm but some of these ideas can be applied to other DRL approaches to the FJSSP and other combinatorial optimization problems. In this section, we present two approaches: the use of DRs to limit the action space and the generation of a diverse set of scheduling policies (DSSP).\\

\textbf{Dispatching Rules to Limit the Action Space.} DRs have been widely used in FJSSP, but their results can be improved, as shown in the next section. However, they serve as a good baseline for the agent's decision-making by limiting the action space. This is particularly relevant on large instances, given the high dimensionality of their action space. In this approach, we consider two simple heuristics to limit the action space: 1) Allowing only the selection of the $k$ operations that can start earlier (if there is a tie, all operations with the same value are allowed) 2) Allowing the $k$ operations that finish earlier. By constraining the action space in this way, we only allow those actions that typically lead to an optimal makespan, as it prevents selecting actions that create large gaps in the scheduling. Moreover, limiting the action space accelerates the model training by reducing the state space that the agent explores. The value $k$ restricts the degree of freedom the agent has in making decisions. We choose these simple heuristics to avoid excessive computational costs that could increase inference time. \\

\textbf{Generating a Diverse Set of Scheduling Policies.} We propose a method for generating a reduced set of solutions generated by a diverse set of models executed in parallel. This strategy is suggested as an alternative to the sampling strategy, as the latter may generate similar solutions since they are generated by the same policy. In summary, the objective of this strategy is to generate a reduced set of policies capable of achieving good results on different types of instances. In other words, we are not interested in generating similar policies; rather, we aim for diverse policies. The method is divided in two steps: 1) Creation of a set of candidate policies for use during inference. 2) Selection of the most representative policies. To find these diverse models, BO has been used to efficiently explore different configurations, and the \textit{K}-Nearest Neighbors algorithm is used for selection. The complete algorithm for creating a DSSP is shown in Algorithm ~\ref{alg:bo}. The following paragraphs will explain the different functions it comprises. \\

\begin{wrapfigure}{r}{0.5\textwidth}
\includegraphics[width=1\linewidth]{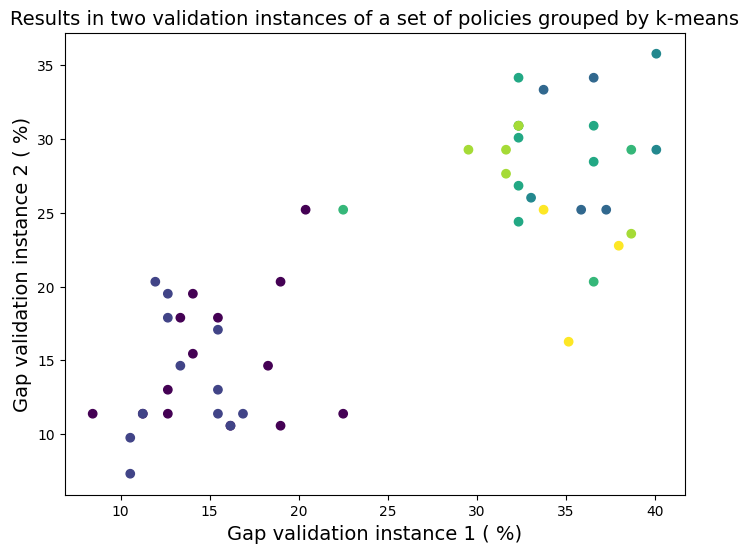} 
\caption{The gap of two instances of the validation set of all the generated candidate policies is shown, colored by the cluster they belong to. All the policies have been divided into 6 clusters. It can be observed that the blue and purple clusters achieve the best results for these two instances.} 
\label{fig:klusters} 
\end{wrapfigure}

Firstly, we define a search space, denoted as $\mathcal{D}$, where we determine which variables to optimize. These variables will be related to the structure of the GNN, the PPO algorithm, the type of DRs used to limit the action space, or the parameters used to generate the instances. Properly defining this domain is crucial to ensuring that the generated policies are diverse. Next, we generate a validation set $\mathcal{V}$ of size $n_{\mathcal{V}}$ (line \ref{bo:line1}) and obtain the optimal (or suboptimal) makespan of each instance using exact methods, stored in the set $\mathcal{VM}$, through the function $\texttt{get\_val\_mk}(\mathcal{V})$. $n_{\mathcal{V}}$ should not be very large to avoid excessive computational cost in the calculation of makespans. For the validation and training of the models, small to medium-sized instances are used, which can be solved without excessive computational cost. We then sample the domain $\mathcal{D}$ using the function $\texttt{sample}(\mathcal{D})$ to obtain hyperparameters $d$. Using $d$, we train a policy using the method described in the previous section, implemented in the function $\texttt{generate\_policy}(d)$.\\

The generated policy is evaluated on $\mathcal{V}$ (line \ref{bo:line5}) and the makespan it produces for each instance is used to calculate the gap compared to the optimal makespan in $\mathcal{VM}$. For a given instance $v \in \mathcal{V}$, the gap for policy $p$ is computed as $g_{p_v} = \frac{(p_v - \mathcal{VM}_v)}{\mathcal{VM}_v}$, where $p_v$ is the makespan achieved by policy $p$ on instance $v$ and $\mathcal{VM}_v$ is the result obtained using exact methods. These gap calculations are implemented in the function $\texttt{eval\_policy}(\mathcal{V}, \mathcal{VM}, p)$. As the first policy is generated, all its results are stored in $\mathcal{CG}$ as the current best gaps achieved. Before starting the BO process, this policy is added to the set of candidate policies to be selected, denoted as $\mathcal{C}$. This initial step is performed so that the next models generated have a baseline result for comparison.\\



The BO process is executed for $n_{\mathcal{BO}}$ iterations (line \ref{bo:line7}). In each iteration $j \in \{ 1,2, \ldots, n_{\mathcal{BO}} \}$, a set of hyperparameters $d_j$ is chosen, and a new policy $p_j$ is generated and evaluated to obtain the gaps $\mathcal{G}_{p_j}$. The maximum difference between the current gaps and those generated by $p_j$ is calculated. This difference serves as the objective function for the BO algorithm, encouraging the generation of policies that achieve good results on validation instances with more room for improvement. If this difference is positive, it means that the new policy has obtained a better result on at least one instance. Therefore, this new policy is added to the set of candidate policies, and the current gaps are updated. \\

Once a set of diverse scheduling policies has been generated, a subset of these policies is selected to run in parallel during inference. The number of policies to parallelize will be determined based on the available computing capacity. To accomplish this, the \textit{K}-Nearest Neighbors algorithm (line \ref{bo:line15}) is applied to the validation set results for each policy to group in $n_\mathcal{P}$ clusters the policies based on similar outcomes in the validation set. The centroid of each cluster is calculated, and the closest policy, using the Euclidean distance, within each cluster is selected as a final candidate, using the function $\texttt{cluster\_center}(\mathcal{K},\mathcal{C})$, and generating the set of policies $\mathcal{P}$ that will be parallelized during inference. This approach ensures that similar policies are not executed during the inference step. In Figure \ref{fig:klusters}, an example can be seen of all the candidate policies and their results on two validation instances, where the $X$ axis shows the gap percentage from the optimal generated by the exact method for the first instance, and the $Y$ axis shows the gap for the second instance. It is observed that the policies of the blue and purple clusters perform better in these instances, while the rest of the clusters do not exhibit such good results. This is because the policies of these clusters are specialized in this type of instances.

\begin{algorithm}
\caption{Algorithm to generate a set of diverse scheduling policies.}
\label{alg:bo} 
\hspace*{\algorithmicindent} \textbf{Input:} BO domain $\mathcal{D}$, BO iterations $n_{BO}$, number of policies $n_\mathcal{P}$, validation set size $n_\mathcal{V}$\\
\hspace*{\algorithmicindent} \textbf{Output:} Policies $\mathcal{P}$
\begin{algorithmic}[1]

\Function{DSSP}{$\mathcal{D}$,$n_\mathcal{P}$, $n_B$, $n_\mathcal{V}$}
    \State Generate a validation set $\mathcal{V}$ \label{bo:line1}
    \State $\mathcal{VM}$ $\gets$ $\texttt{get\_val\_mk}(\mathcal{V})$ \label{bo:line2}
    \State $d \gets$ $\texttt{sample}(\mathcal{D})$ \label{bo:line3}
    \State $p \gets$ $\texttt{generate\_policy}(d)$ \label{bo:line4}
    \State $\mathcal{CG}$ $\gets$ $\texttt{eval\_policy}(\mathcal{V}$, $\mathcal{VM}$, p) \label{bo:line5}
    \State $\mathcal{C}$ $\gets$ $\{p\}$ \label{bo:line6}
    \For{$j=1$ to $n_{BO}$} \label{bo:line7}
        \State $d_j$ $\gets$ \texttt{sample}($\mathcal{D}$) \label{bo:line8}
        \State $p_j$ $\gets$ \texttt{generate\_policy}($d_j$) \label{bo:line9}
        \State $\mathcal{G}_{p_j}$ $\gets$ \texttt{eval\_policy}($\mathcal{V}$, $\mathcal{VM}$, $p_j$ ) \label{bo:line10}
        \State $g_j \gets \text{max}(\mathcal{CG} - \mathcal{G}_{p_j})$ \label{bo:line11}
        \If{$g_j > 0$} \label{bo:line12}
            \State $\mathcal{C}$ $\gets$ $\mathcal{P} \cup \{p_j\}$ \label{bo:line13}
            \State $\mathcal{CG}_v$ $\gets$ $\min\limits_{v \in \mathcal{V}}(\mathcal{CG}_v, \mathcal{G}_{p_{j_v}})$  \label{bo:line14}
        \EndIf
        Update surrogate model with $(d_j, g_j)$\label{bo:line15}
    \EndFor

    \State $\mathcal{K} \gets \texttt{KNN}(\mathcal{C}, n_\mathcal{P})$ \label{bo:line16}
    \State $\mathcal{P} \gets \texttt{cluster\_center}(\mathcal{K},\mathcal{C})$ \label{bo:line17}

    \Return $\mathcal{P}$
       \EndFunction

\end{algorithmic}
\end{algorithm}

\section{Experimental results}\label{sec:experimentalresult}
In this section, we present the experiments conducted to evaluate the validity of the proposed methodology, which we refer to as Enhanced Diverse Scheduling Policies (EDSP), we study its capability to outperform DRs and its competitiveness with three state-of-the-art DRL-based approaches. For this purpose, comparisons have been made with two publicly available FJSSP benchmarks\footnote{We obtained the benchmarks from \url{https://openhsu.ub.hsu-hh.de/handle/10.24405/436}.}, measuring both the quality of the solutions and the time to generate them. 

\subsection{Generation of Synthetic Instances}

To train the models, synthetic instances must be generated, and for this purpose, the instance generation method proposed in \cite{brandimarte1993routing}, also used by \cite{song2022flexible}, has been adapted. The generation of synthetic instances is governed by various parameters that determine their size and structure. These parameters include the minimum and maximum number of jobs ($j_{\text{min}}$ and $j_{\text{max}}$), the minimum and maximum number of machines ($m_{\text{min}}$ and $m_{\text{max}}$), the minimum and maximum number of operations per job ($o_{\text{min}}$ and $o_{\text{max}}$), the maximum number of options each operation can have ($op_{\text{max}}$, with a minimum fixed at four), the maximum processing time for an operation $\overline{p}$ (with a minimum defined as one), and the deviation from the mean $d$. To generate an instance, the uniform distribution within the defined ranges is sampled. The processing time of the operation $o_{ij}$ on machine $m_k$, if is compatible,  is calculated by sampling from the uniform distribution $\mathcal{U}(\overline{p_{ij}}(1-d), \overline{p_{ij}}(1+d))$.  

\subsection{Configuration}

For the experiments, the number of BO iterations $n_{\text{BO}}$ has been set to 50 and the tree-structured Parzen Estimator has been used as the sampler. The validation set size, $n_{\mathcal{V}}$, has been set to 100 and have been generated by randomly selecting from the number of jobs, machines and operation ranges defined below. The size of the policy set $n_{\mathcal{P}}$ has been set to 6, to enable efficient parallelization of the models and to achieve similar execution times to the other DRL methods, in order to compare them fairly. Regarding the domain of BO, the hyperparameters can be divided into categories related to: the structure of the GNN, the training mode of the PPO algorithm, the size of the generated instances, and the DR and degrees of freedom for decision-making. The following ranges have been considered:
\begin{itemize}
\item The defined ranges for generating instance sets in each iteration of BO were: $j_{\text{min}} \in [5,9]$, $j_{\text{max}} \in [10,15]$, $m_{\text{min}} \in [4,7]$, $m_{\text{max}} \in [9,13]$, $op_{\text{max}} \in [5,9]$, $p \in [8,25]$. $d$ has been set to $0.2$.
\item Related to the GNN: the number of layers $L \in [1,2]$ and the number of hidden channels $d^\prime \in \{32, 64, 128\}$. For simplicity, the dimension of the node embeddings has been considered the same for all nodes. $MLP_{1\theta}$ and $MLP_{2\theta}$ of the actor and the critic have a single layer of the same dimension.
\item Related to network training: the maximum number of episodes during training $ n_{eps} \in [10000,15000]$, learning rate $lr \in [0.001,0.0001]$, batch size $b \in \{64, 128, 256\}$, and the model training frequency $n_t \in [10,100]$.
\item The DR used by the policy, as well as the possible options $k \in [1,3]$.
\end{itemize}
Allowing the BO algorithm to have many hyperparameter options aims to obtain diverse models that can be effective on different types of instances. For the PPO algorithm, the discount factor $\gamma$ was fixed to 1, the clipping parameter $\epsilon$ to 0.2, and the coefficients for policy, value, and entropy in the loss function were set to 1, 0.5, and 0.01, respectively. The number of epochs per episode was set to 3, and Adam optimizer \citep{kingma2014adam} was used.\\

The method was implemented using Python 3.10, PyTorch Geometric \citep{Fey/Lenssen/2019} was utilized for the GNNs, Optuna \citep{akiba2019optuna} was used for hyperparameter optimization and OR-Tools for the exact method \footnote{ We obtained the implementation from \url{https://github.com/google/or-tools/blob/stable/examples/python/flexible_job_shop_sat.py}.}. The hardware is a machine with AMD Ryzen 5 5600X processor and an NVIDIA GeForce RTX 3070 Ti. The code will be made publicly available upon acceptance of the paper.

\subsection{Results in publicly available FJSSP benchmarks}
To evaluate the quality of the solutions generated by EDSP, the makespan has been chosen as the objective to minimize, the same as the other methods. The metric with which different approaches are compared is the optimal gap. For method $m$ and instance $i$, this gap is defined as:
$$
g_{m_i} = \frac{(C_{max}^{m_i} - C_{max}^{o_i})}{C_{max}^{o_i}}
$$
where $C_{max}^{m_i}$ is the makespan generated by method $m$ for instance $i$, and $C_{max}^{o_i}$ is the best solution or the approximate optimal solution for that instance.\\

In relation to the selected benchmarks, 45 large instances from the Behnke benchmark \citep{behnke2012test} have been used and, although our main focus is large instances, we have also used the instances proposed by  \cite{hurink1994tabu} (40 small and medium-sized instances from the vdata set) to assess that our method still is a competitive alternative for small-medium ones. We have also chosen these subsets from these benchmarks in order to compare our results with the approach presented by \cite{lei2022multi}, as they use this benchmark. In addition to comparing it with this approach, we compare our method with the ones proposed by \cite{wang2023flexible} and \cite{song2022flexible}, both also recent DRL methods, using their publicly available open-source code and using their default settings to train and test the models. As far as we know, these methods represent the state-of-the-art of DRL-based methods to solve the FJSSP.\\

\begin{figure}[!h] 
	\centering 
 \includegraphics[width=1\columnwidth]{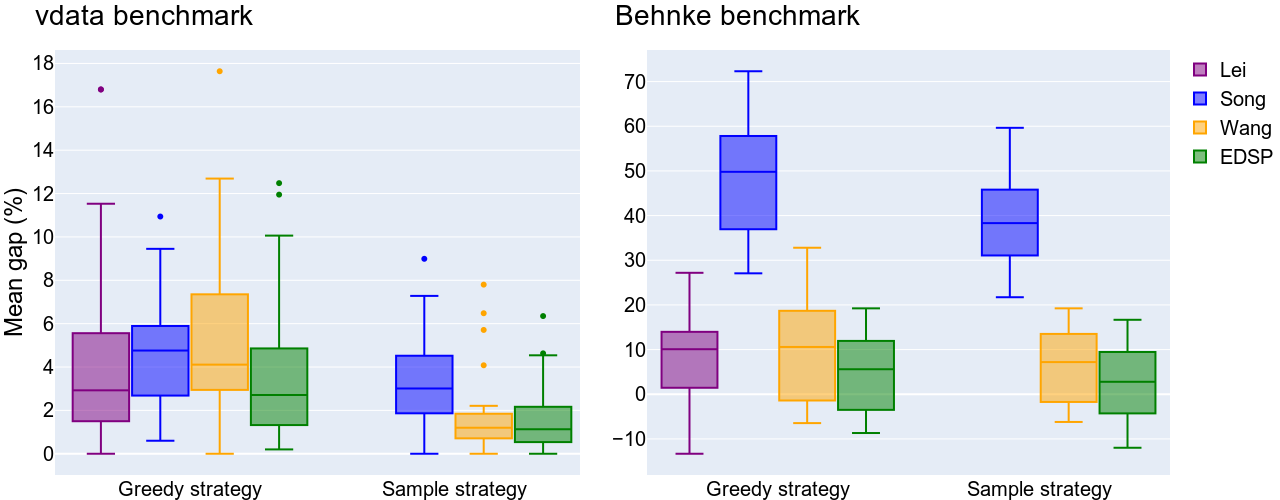}
	\caption{Comparison of DRL-based methods on the vdata and Behnke benchmarks using different strategies.} 
	\label{fig:benchmarks} 
\end{figure}

As mentioned earlier, in the literature there are two methods for generating solutions using a policy: the greedy strategy and the sampling strategy. The greedy strategy involves selecting the action with the highest probability from $\pi_\theta$, generating a single solution, while the sampling strategy involves sampling from the distribution $\pi_\theta$ instead of selecting the maximum probability and choosing the solution with the smallest makespan value among all runs, albeit at the expense of increased execution time. In the case of \cite{song2022flexible} and \cite{wang2023flexible} this is done by parallelizing a instance 100 times. However, the authors of \cite{lei2022multi} do not employ a sampling strategy, only a greedy one. The greedy strategy has been compared with the use of a single policy from the EDSP, selecting the one that achieves the best results on the validation set, to ensure a fair comparison between the different approaches. This comparison is considered fair since the execution times of the different approaches using the greedy strategy are similar, and only one solution is generated. In Figure \ref{fig:benchmarks}, a general overview of the results of the four methods in the two benchmarks using both policies is shown, showing that our method outperforms the other approaches. In the next subsection, a more detailed analysis of the results will be conducted.   \\

Additionally, in each benchmark, we have also compared the DRL methods with DRs. To define a DR for solving the FJSSP, two rules need to be defined: one for job selection and sequencing, and another for operation assignment to machines. Regarding the former, the following rules have been considered:

\begin{itemize}
    \item First In First Out (FIFO), which processes the first operation to arrive in the queue of a machine.
    \item Most Work Remaining (MWKR), which processes a job with the highest remaining workload.
    \item Least Work Remaining (LWKR), which processes a job with the lowest remaining workload.
    \item Most Operation Number Remaining (MOPNR), which processes a job with the highest number of remaining operations.
\end{itemize}

For the machine selection criteria, the following rules have been considered:

\begin{itemize}
    \item Shortest Processing Time (SPT), which assigns a machine with the shortest processing time for an operation.
    \item Earliest Ending Time (EET), which selects the machine where an operation can start earliest.
\end{itemize}

Those DRS have been shown to be effective in the literature \citep{jun2019learning} and have recently been used to compare them with other DRL-based methods \citep{lei2022multi, song2022flexible}.\\

\textbf{Results in vdata benchmark.} Table \ref{tab:vdataresults} presents the results obtained by the four DRL methods and the best DR, grouped according to the instance sizes. The gap is calculated by comparing the solutions achieved by each approach with the best upper bounds reported in the existing literature, as published by \cite{behnke2012test}. Table \ref{tab:vdataresults} highlights the best gaps and the shortest execution times obtained by the approaches.  \\

As expected, the results of the best DR, which is FIFO for job selection and EET for machine selection, yield the worst results, with a mean gap of $7.94\%$. Concerning the greedy strategy, the mean gaps are $4.35\%$ for \cite{song2022flexible}, $5.5\%$ for \cite{wang2023flexible} and $4.20\%$ for \cite{lei2022multi}. Our method achieves the best result with a mean gap of $3.47\%$. For the sampling strategy, the best mean gap is achieved by \cite{wang2023flexible}, with a gap of $1.81\%$. Our EDSP method closely follows with the second-best results, featuring an average gap of $1.94\%$, nearly identical to the previous. This is followed by the method of \cite{song2022flexible} at $3.72\%$. We emphasize that, as seen in Table \ref{tab:vdataresults}, the average results for larger instances are better with our method, suggesting that our approach is better with this kind of instances. For the sampling strategy, we speculate that the better outcomes in smaller instances by \cite{wang2023flexible} may be attributed to their state representation featuring more features and their GNN utilizing a dual attention heads architecture. As for the average execution times, which are shown in Table \ref{tab:vdataresults}, we compared the proposed methods except for the one by \cite{lei2022multi} as they did not provide their exact times. As can be seen in Table \ref{tab:vdataresults}, the DRL-based methods achieve similar execution times.

\begin{table}[H]
\centering
\caption{Comparison of the mean gap grouped by size and strategy obtained by DRs, the methods proposed by \cite{song2022flexible}, \cite{wang2023flexible}, \cite{lei2022multi} and our approach (EDSP) in vdata instances.}
\label{tab:vdataresults}
\resizebox{\textwidth}{!}{
\begin{tabular}{c|c||r|r|r|r|r||r|r|r}

\multirow{2}{*}{Size} &  \multirow{2}{*}{Obj} & \multirow{2}{*}{Best DR} &\multicolumn{4}{c||}{Greedy strategy} & \multicolumn{3}{c}{Sampling strategy}\\
& & &  Song & Wang &  Lei & EDSP  & Song & Wang  & EDSP \\

\hline
\hline

 10x5 & Gap(\%) &  14.30 &   \textbf{6.98} & 8.55 & 8.00 & 7.12 & 6.89 &      \textbf{3.09} &         4.14 \\
      & Time (s.) & \textbf{0.03} &  0.54 &  0.44 & - &  0.62 &  4.79 &  0.72 &  2.67 \\
\hline
 10x10 & Gap(\%) &  4.76 & 3.84 & 5.03 & \textbf{0.32} & 3.82  &      0.28 &      \textbf{0.27} &        1.24 \\
      & Time (s.) & \textbf{0.06} &  1.11 &  0.86 & - &  0.81 &  9.10 &  1.75 &  2.77 \\
\hline    
 15x5 & Gap(\%) &  6.96 & 4.03 & 5.53 & 3.24 & \textbf{3.02} &      2.99 &      \textbf{1.13} &         2.10 \\
      & Time (s.) & \textbf{0.04} &  0.84&  0.65 & - &  0.91 &  6.38 &  1.05 &   2.38 \\
\hline    
 15x10 & Gap(\%) &  15.66 & \textbf{5.91 }& 11.12 & 11.44 & 6.41 & 7.35 &      5.13 &       \textbf{4.07}  \\
      & Time (s.) & \textbf{0.10} &  1.69&  1.35 & - &  1.73 & 13.19 &  3.47 &   4.17  \\  
\hline    
 15x15 & Gap(\%) &  4.31 & 3.93 & 3.58 & 3.25 & \textbf{2.51}  &      \textbf{0.88} &      \textbf{0.88} &         0.91 \\
      & Time (s.) & \textbf{0.16} &  2.49&  2.04 & - &  2.92 & 24.11 &  7.85 &  7.52  \\ 
\hline    
 20x5 & Gap(\%) & 4.51 & 3.49 & 2.74 & 1.85 & \textbf{1.58} & 2.97 &      0.94 &        \textbf{0.88}  \\
      & Time (s.) & \textbf{0.07} &  1.10&  0.91 & - &  1.14 &  7.90 &  1.58 &   2.92 \\ 
\hline    
 20x10 & Gap(\%) &  7.62 & 3.95 & 4.25 & 4.08 & \textbf{2.39} & 5.02 &      1.79 &         \textbf{1.50}  \\
      & Time (s.) & \textbf{0.14} &  2.21&  1.75 & - &  2.51  & 18.14 &  5.29 &    6.11 \\ 
\hline    
 30x10 & Gap(\%) &  5.38 & 2.63 & 3.17 & 2.00 & \textbf{0.91} &      3.25 &      1.16 &      \textbf{0.57} \\
      & Time (s.) & \textbf{0.27} &  3.24&  2.70 & - &  3.88 & 30.36 & 11.48 &   12.10  \\ 
\hline


\end{tabular}}
\end{table}

\textbf{Results in Behnke benchmark.} Table \ref{tab:behnkeresults} presents the results categorized by instance size from the Behnke benchmark, including the outcomes obtained by the best DR (FIFO for job selection and SPT for machine selection), and the four DRL methods. As mentioned, this benchmark is particularly relevant to this paper as it consists of instances larger than the previous one. In this scenario, no optimal solutions are available, so the gap is calculated based on the solutions generated using OR-Tools. OR-Tools was executed for 1800 seconds, employing the CP-SAT solver, and has been widely utilized in various papers for this purpose \citep{song2022flexible, wang2023flexible, tassel2021reinforcement}, because it is one of the best open-source tools for solving scheduling problems \citep{da2019industrial}.  In \ref{appendix:table}, the results for each of the instances are displayed. It is important to show the results obtained by the solver on each instance because, for example, in comparison, \cite{lei2022multi} use the Gurobi solver for 3600 seconds and obtain considerably worse results than OR-Tools.\\

In this case, our approach clearly outperforms the other DRL-based methods, both using the sampling strategy and the greedy strategy. For the second strategy, the best mean gap is obtained with our approach at $4.98\%$, followed by \cite{lei2022multi} with $7.27\%$ and \cite{wang2023flexible} with $10.56\%$. There are significant differences compared to the approach of \cite{song2022flexible}, which achieves a mean gap of $48.6\%$—even worse than the best DR, which has a mean gap of $27.91\%$. We speculate that this is because, unlike other methods, in this approach, the output of the GNN is a single vector representing the state. This vector is utilized by the policy to predict the probability of all actions. However, in larger instances, a single vector may be insufficient to encapsulate all the information of a state. Regarding the sampling strategy, our approach is always better than the other two approaches, except in two groups of instances where the greedy strategy of \cite{lei2022multi} outperforms. Furthermore, as observed in Figure \ref{fig:benchmarks}, the variability of our method is reduced compared to other approaches. \\

Regarding execution time, as shown in Table \ref{tab:behnkeresults}, mainly in the sampling strategy it can be observed that as the instance dimension increases, the processing time significantly increases for \cite{wang2023flexible}, which is slower than our approach. This could be because they calculate more features, particularly within the operation-type nodes, which are more numerous.




\begin{table}[H]
\centering
\caption{Comparison of the mean gap grouped by size and strategy obtained by DRs, the methods proposed by \cite{song2022flexible}, \cite{wang2023flexible}, \cite{lei2022multi} and our approach in the Behnke benchmark.}
\label{tab:behnkeresults}
\resizebox{\textwidth}{!}{
\begin{tabular}{c|c||r|r|r|r|r||r|r|r}

\multirow{2}{*}{Size} &  \multirow{2}{*}{Obj} & \multirow{2}{*}{Best DR} &\multicolumn{4}{c||}{Greedy strategy} & \multicolumn{3}{c}{Sampling strategy}\\
& & &  Song & Wang &  Lei & EDSP  & Song & Wang  & EDSP \\

\hline
\hline

 20x20 & Gap(\%) &  42.38 & 59.69 & 28.24 & \textbf{11.93} & 14.87 & 42.65 &     17.55 &         \textbf{10.53} \\
      & Time (s.) & \textbf{0.08} &  1.23 & 0.90 & - &  1.74 &   10.29 &   2.78 & 3.60 \\
\hline
 20x40 & Gap(\%) &   42.11 & 59.85 & 18.42 & 16.19 & \textbf{14.08}  &      50.57 &     13.86 &        \textbf{12.53} \\
      & Time (s.) & \textbf{0.09} &  1.11 & 1.00& - &  1.92 &  11.39 &   6.43 &  6.67 \\
\hline    
 20x60 & Gap(\%) & 45.47 & 60.81 & 25.78 & 20.23 & \textbf{14.46} &       50.22 &     14.36 &         \textbf{12.64} \\
      & Time (s.) & \textbf{0.10} &  1.18 & 1.06 & - &  2.19 &  14.38 &  12.35 &  12.06 \\
\hline    
 50x20 & Gap(\%) &  26.72 & 46.48 & 11.51 & \textbf{0.01} & 3.93 &      34.44 &      8.11 &       1.94  \\
      & Time (s.) & \textbf{0.34} &  2.82 & 2.44 & - &  3.16 &  29.16 &  14.64 &   11.88  \\  
\hline    
 50x40 & Gap(\%) &  25.26 & 50.95 & 11.96 & 11.88 & \textbf{5.09}  &       40.59 &      7.15  &      \textbf{3.27}  \\
      & Time (s.) & \textbf{0.37} &  2.49&  2.04 & - &  2.94 & 30.67 &  54.79 &  26.07  \\ 
\hline    
 50x60 & Gap(\%) & 27.33 & 55.10 & 7.88 & 12.27 & \textbf{6.41} & 44.09 &      6.13 &        \textbf{3.78}  \\
      & Time (s.) & \textbf{0.39} &  2.88 & 2.85 & - &  3.40 &  37.83 &  65.82 &    41.53 \\ 
\hline    
100x20 & Gap(\%) &  11.84 & 30.11 & -3.65 & \textbf{-11.02} & -7.54 &  23.70 &     -4.73 &         -9.36  \\
      & Time (s.) & \textbf{1.13} &  5.51 & 5.08 & - &  5.88 &  68.88 &  54.79 &    61.05\\ 
\hline    
 100x40 & Gap(\%) &  16.76 & 36.72 & \textbf{-3.95} & 2.03 & -2.84 &      28.21 &     -2.69 &      \textbf{-4.95} \\
      & Time (s.) & \textbf{1.21} &  5.43 & 5.65 & - &  6.16 & 82.28 & 134.13 &   93.44  \\ 
\hline

 100x60 & Gap(\%) &  14.30 & 37.64 & -1.15 & 1.92 & \textbf{-3.66}  &      31.85 &     -1.92 &      \textbf{-4.21} \\
      & Time (s.) & \textbf{1.29} &  5.52 & 7.88 & - &  7.67 & 102.07 & 235.10 &   130.15  \\ 
\hline


\end{tabular}}
\end{table}

\section{Conclusions and future work}\label{sec:conclusion}
This paper has introduced a novel representation of the FJSSP as an MDP and a new approach based on DRL and GNNs to dynamically solve the FJSSP. To achieve this, we first proposed a new representation of states, action space, and transition function, where a job-type node is proposed and unnecessary nodes and edges are progressively removed to make our method more efficient. Moreover, as part of our approach we presented two new strategies to enhance the performance of the DRL method. This was achieved by creating a set of diverse policies that can be executed in parallel and by constraining the action space using DRs. \\

To validate our approach, we compared it with two public benchmarks and three state-of-the-art DRL-based approaches, using two different strategies. Except for the sampling strategy in one of the benchmarks with one of the methods, where the results were slightly lower, our approach proves superior to the other DRL methods in small to medium-sized instances. Furthermore, in large instances, it significantly outperformed all three methods. Notably, our method even outperformed OR-Tools on instances with a higher number of jobs and machines, all while maintaining significantly lower execution times.\\

For future work, we plan to extend this method to handle additional complexities in the FJSSP, such as new job insertions or setup times. Moreover, we aim to apply our approach to the low-carbon FJSSP, which considers both the makespan and energy consumption of machines. Additionally, it will be interesting to study the possibility to extend this approach to other combinatorial optimization and operations research problems, such as hub location problems or knapsack problems.


\section*{Acknowledgements}
This work was partially financed by the Basque Government through their Elkartek program (SONETO project, ref. KK-2023/00038) and the Gipuzkoa Provicial Council through their Gipuzkoa network of Science, Technology and Innovation program (KATEAN  project, ref. 2023-CIEN-000053-01). R. Santana thanks the Misiones Euskampus 2.0 programme for the financial help received through Euskampus Fundazioa. Partial support by the Research Groups 2022-2024 (IT1504-22) and the Elkartek Program (KK-2020/00049, KK-2022/00106, SIGZE, K-2021/00065) from the Basque Government, and the PID2019-104966GB-I00 and PID2022-137442NB-I00 research projects from the Spanish Ministry of Science is acknowledged.

\bibliographystyle{elsarticle-harv} 
\bibliography{main}

\newpage

\appendix
\section{Public benchmarks results}\label{appendix:table}
In the following table, the results obtained by the different methods are shown.

\setcounter{table}{0}

\begin{table}[H]
\centering
\small
\caption{Comparison of the mean gap grouped by size and strategy obtained by DRs, the methods proposed by \cite{song2022flexible}, \cite{wang2023flexible}, \cite{lei2022multi} and  our approach (EDSP) in vdata instances. The upper bounds are obtained from \cite{behnke2012test}.}
\label{tab:resultsallvdata}
\resizebox*{\textwidth}{\dimexpr\textheight-10\baselineskip\relax}{
\begin{tabularx}{\textwidth}{c|c||X|X|X|X|X||X|X|X||X}

\multirow{2}{*}{Index} &  \multirow{2}{*}{Size}  & \multirow{2}{*}{DR} & \multicolumn{4}{c||}{Greedy strategy} & \multicolumn{3}{c||}{Sampling strategy} & \multirow{2}{*}{UB}\\
& & &  Song & Wang &  Lei & EDSP  & Song & Wang  & EDSP &  \\

\hline
\hline

la01 & 10x5 & 660 & \textbf{579} & 620 & 610 & 599 & 607 & 583 & \textbf{580} & 570 \\
la02 & 10x5 & 584 & 579 & 572 & \textbf{555} & 559 & 553 & \textbf{539} & 551 & 529 \\
la03 & 10x5 & 551 & \textbf{513} & 525 & 532 & 525 & 506 & \textbf{493} & 506 & 477 \\
la04 & 10x5 & 600 & \textbf{529} & 545 & 530 & 562 & 544 & \textbf{522} & 526 & 502 \\
la05 & 10x5 & 504 & 507 & 490 & 507 & \textbf{470} & 498 & \textbf{475} & \textbf{475} & 457 \\
la06 & 15x5 & 839 & 823 & 831 & \textbf{820} & 832 & 827 & \textbf{808} & \textbf{808} & 799 \\
la07 & 15x5 & 801 & 785 & 828 & \textbf{757} & 792 & 781 & 759 & 775 & 749 \\
la08 & 15x5 & 808 & 806 & 810 & 782 & \textbf{779} & 782 & \textbf{775} & 776 & 765 \\
la09 & 15x5 & 910 & 871 & 886 & 879 & \textbf{867} & 876 & \textbf{854} & 874 & 853 \\
la10 & 15x5 & 889 & 843 & 831 & 862 & \textbf{818} & 822 & \textbf{818} & 820 & 804 \\
la11 & 20x5 & 1137 & 1122 & 1120 & \textbf{1101} & 1103 & 1099 & \textbf{1080} & 1086 & 1071 \\
la12 & 20x5 & 965 & \textbf{950} & 959 & \textbf{950} & 951 & 968 & \textbf{944} & \textbf{944} & 936 \\
la13 & 20x5 & 1072 & 1065 & 1063 & 1053 & \textbf{1050} & 1068 & 1045 & \textbf{1042} & 1038 \\
la14 & 20x5 & 1137 & 1108 & 1099 & \textbf{1086} & 1087 & 1100 & 1086 & \textbf{1085} & 1070 \\
la15 & 20x5 & 1130 & 1144 & 1106 & 1111 & \textbf{1095} & 1123 & 1098 & \textbf{1093} & 1089 \\
la16 & 10x10 & 727 & 764 & 752 & \textbf{717} & 735 & 719 & 723 & 721 & 717 \\
la17 & 10x10 & 708 & 659 & 728 & \textbf{647} & 650 & \textbf{646} & \textbf{646} & 653 & 646 \\
la18 & 10x10 & 687 & 667 & \textbf{663} & \textbf{663} & 674 & \textbf{663} & \textbf{663} & \textbf{663} & 663 \\
la19 & 10x10 & 637 & 660 & 641 & \textbf{626} & 694 & 624 & \textbf{620} & 645 & 617 \\
la20 & 10x10 & 801 & 779 & 784 & \textbf{756} & 770 & \textbf{756} & \textbf{756} & \textbf{756} & 756 \\
la21 & 15x10 & 895 & \textbf{830} & 947 & 887 & 834 & 855 & 851 & \textbf{826} & 805 \\
la22 & 15x10 & 843 & 787 & 816 & 793 & \textbf{782} & 782 & \textbf{765} & 769 & 735 \\
la23 & 15x10 & 906 & 849 & 876 & 858 & \textbf{848} & 849 & \textbf{826} & 842 & 813 \\
la24 & 15x10 & 919 & \textbf{816} & 826 & 883 & 817 & 831 & 805 & \textbf{804} & 756 \\
la25 & 15x10 & 903 & \textbf{809} & 831 & 883 & 829 & 830 & 815 & \textbf{780} & 756 \\
la26 & 20x10 & 1166 & 1079 & 1105 & 1089 & \textbf{1063} & 1111 & 1069 & \textbf{1061} & 1053 \\
la27 & 20x10 & 1170 & 1136 & 1137 & 1123 & \textbf{1112} & 1135 & 1104 & \textbf{1095} & 1084 \\
la28 & 20x10 & 1133 & 1124 & 1101 & 1106 & \textbf{1098} & 1118 & \textbf{1090} & 1092 & 1069 \\
la29 & 20x10 & 1058 & 1029 & 1038 & 1049 & \textbf{1007} & 1051 & 1017 & \textbf{1010} & 995 \\
la30 & 20x10 & 1145 & \textbf{1111} & 1113 & 1117 & 1117 & 1119 & \textbf{1084} & 1091 & 1069 \\
la31 & 30x10 & 1623 & 1547 & 1586 & 1561 & \textbf{1527} & 1598 & 1537 & \textbf{1531} & 1524 \\
la32 & 30x10 & 1798 & 1706 & 1687 & 1693 & \textbf{1676} & 1704 & 1677 & \textbf{1673} & 1661 \\
la33 & 30x10 & 1567 & 1533 & 1539 & 1531 & \textbf{1511} & 1536 & 1515 & \textbf{1504} & 1497 \\
la34 & 30x10 & 1589 & 1579 & 1583 & 1562 & \textbf{1553} & 1580 & 1551 & \textbf{1545} & 1535 \\
la35 & 30x10 & 1601 & 1606 & 1615 & 1574 & \textbf{1570} & 1607 & 1583 & \textbf{1565} & 1549  \\
la36 & 15x15 & 1003 & 961 & 987 & 985 & \textbf{950} & 955 & \textbf{952} & 953 & 948 \\
la37 & 15x15 & \textbf{1013} & 1041 & 1055 & 1028 & 1018 & 1003 & 1004 & \textbf{999} & 986 \\
la38 & 15x15 & 990 & 996 & \textbf{946} & 948 & \textbf{946} & \textbf{943} & \textbf{943} & 947 & 943 \\
la39 & 15x15 & 994 & \textbf{936} & 976 & 979 & 973 & \textbf{940} & 942 & 942 & 922 \\
la40 & 15x15 & \textbf{957} & 1008 & 961 & 968 & 986 & \textbf{955} & \textbf{955} & 956 & 955 \\

\hline
\end{tabularx}}

\end{table}
\begin{table}[h]
\small
\centering
\caption{Comparison of the mean gap grouped by size and strategy obtained by DRs, the methods proposed by \cite{song2022flexible}, \cite{wang2023flexible}, \cite{lei2022multi} and our approach (EDSP) and OR-Tools (with a time limit of 1800 seconds) in Behnke instances.}
\label{tab:resultsallbehnke}
\resizebox*{\textwidth}{\dimexpr\textheight-5\baselineskip\relax}{
\begin{tabularx}{\textwidth}{c|c||X|X|X|X|X||X|X|X||X}
\multirow{2}{*}{Index} &  \multirow{2}{*}{Size}  & \multirow{2}{*}{DR} & \multicolumn{4}{c||}{Greedy strategy} & \multicolumn{3}{c||}{Sampling strategy} & \multirow{2}{*}{Opt}\\
& & &  Song & Wang &  Lei & EDSP  & Song & Wang  & EDSP &  \\

\hline
\hline

06 & 20x20 & 176 & 195 & 166 & 143 & \textbf{141} & 179 & 148 & \textbf{141} & 128 \\
07 & 20x20 & 174 & 199 & 162 & \textbf{142} & 151 & 176 & 152 & \textbf{138} & 129 \\
08 & 20x20 & 208 & 198 & 160 & 139 & \textbf{138} & 185 & 148 & 142 & 126 \\
09 & 20x20 & 169 & 203 & 166 & \textbf{144} & 148 & 180 & 147 & \textbf{142} & 125 \\
10 & 20x20 & 181 & 224 & 164 & \textbf{146} & 155 & 190 & 155 & \textbf{142} & 130 \\
11 & 50x20 & 321 & 346 & 291 & \textbf{250} & 257 & 337 & 276 & 251 & 257 \\
12 & 50x20 & 282 & 367 & 276 & \textbf{247} & 250 & 336 & 267 & 250 & 251 \\
13 & 50x20 & 315 & 367 & 272 & \textbf{249} & 273 & 343 & 271 & 265 & 248  \\
14 & 50x20 & 352 & 386 & 282 & \textbf{257} & 258 & 342 & 275 & 258 & 251 \\
15 & 50x20 & 324 & 376 & 282 & \textbf{255} & 269 & 333 & 271 & 258 & 251 \\
16 & 100x20 & 547 & 637 & 464 & \textbf{437} & 449 & 598 & 460 & 444 & 486 \\
17 & 100x20 & 562 & 606 & 451 & \textbf{430} & 447 & 597 & 449 & 439 & 475 \\
18 & 100x20 & 524 & 642 & 476 & \textbf{428} & 444 & 600 & 466 & 440 & 478 \\
19 & 100x20 & 539 & 637 & 473 & \textbf{423} & 447 & 594 & 468 & 436 & 488 \\
20 & 100x20 & 524 & 615 & 459 & \textbf{427} & 442 & 593 & 454 & \textbf{426} & 484 \\
26 & 20x40 & 167 & 187 & 135 & 129 & \textbf{128} & 173 & 132 & \textbf{126} & 113 \\
27 & 20x40 & 180 & 194 & 145 & \textbf{140} & 141 & 180 & \textbf{137} & 140 & 123 \\
28 & 20x40 & 152 & 181 & 135 & 133 & \textbf{125} & 182 & 132 & \textbf{121} & 114 \\
29 & 20x40 & 145 & 185 & 140 & 138 & \textbf{134} & 177 & \textbf{131} & 134 & 117 \\
30 & 20x40 & 185 & 191 & \textbf{140} & 142 & 142 & 171 & \textbf{136} & 140 & 120 \\
31 & 50x40 & 308 & 353 & 261 & 271 & \textbf{250} & 326 & 253 & \textbf{242} & 236 \\
32 & 50x40 & 287 & 346 & 267 & 267 & \textbf{241} & 326 & 254 & \textbf{241} & 235 \\
33 & 50x40 & 274 & 340 & 261 & 261 & \textbf{248} & 314 & 248 & \textbf{247} & 234 \\
34 & 50x40 & 280 & 356 & 268 & 250 & \textbf{245} & 338 & 250 & \textbf{239} & 232 \\
35 & 50x40 & 303 & 355 & 242 & 249 & \textbf{235} & 326 & 238 & \textbf{229} & 223 \\
36 & 100x40 & 500 & 633 & 422 & 442 & \textbf{421} & 575 & 429 & \textbf{414} & 432 \\
37 & 100x40 & 504 & 599 & \textbf{420} & 444 & 431 & 567 & 428 & \textbf{423} & 449 \\
38 & 100x40 & 490 & 593 & \textbf{424} & 454 & 429 & 570 & 427 & \textbf{422} & 445  \\
39 & 100x40 & 498 & 598 & 429 & 471 & \textbf{423} & 574 & 432 & \textbf{412} & 447 \\
40 & 100x40 & 538 & 619 & \textbf{443} & 460 & 459 & 567 & 450 & \textbf{445} & 453 \\
46 & 20x60 & 176 & 194 & 142 & 145 & \textbf{131} & 172 & \textbf{130} & 131 & 114 \\
47 & 20x60 & 165 & 187 & 145 & 144 & \textbf{127} & 177 & 131 & \textbf{127} & 117 \\
48 & 20x60 & 165 & 195 & 147 & \textbf{141} & 146 & 182 & 143 & 142 & 125 \\
49 & 20x60 & 157 & 194 & 150 & 134 & \textbf{129} & 171 & 131 & \textbf{126} & 113 \\
50 & 20x60 & 198 & 180 & 160 & 147 & \textbf{145} & 187 & 142 & \textbf{141} & 123 \\
51 & 50x60 & 283 & 342 & \textbf{244} & 253 & 252 & 328 & \textbf{238} & 240 & 226 \\
52 & 50x60 & 287 & 354 & 237 & 242 & \textbf{234} & 323 & 236 & \textbf{233} & 226 \\
53 & 50x60 & 280 & 358 & 239 & 256 & \textbf{235} & 323 & 236 & \textbf{231} & 222 \\
54 & 50x60 & 302 & 360 & 256 & 268 & \textbf{251} & 343 & 251 & \textbf{243} & 237 \\
55 & 50x60 & 301 & 355 & 255 & 262 & \textbf{242} & 327 & 250 & \textbf{237} & 230 \\
56 & 100x60 & 505 & 611 & 424 & 439 & \textbf{419} & 569 & 425 & \textbf{411} & 434 \\
57 & 100x60 & 526 & 592 & 430 & 442 & \textbf{414} & 565 & 427 & \textbf{412} & 432 \\
58 & 100x60 & 475 & 624 & 443 & 442 & \textbf{424} & 592 & 428 & \textbf{424} & 440 \\
59 & 100x60 & 458 & 597 & \textbf{422} & 443 & 426 & 567 & 426 & \textbf{424} & 433 \\
60 & 100x60 & 530 & 579 & 438 & 458 & \textbf{419} & 584 & 434 & \textbf{419} & 443 \\

\hline


\end{tabularx}}
\end{table}

\end{document}